\documentclass[11pt]{article}
\usepackage{coling2016}
\usepackage{times}
\usepackage{url}
\usepackage[dvipsnames]{xcolor}
\makeatletter
\newcommand{\@BIBLABEL}{\@emptybiblabel}
\newcommand{\@emptybiblabel}[1]{}
\makeatother
\usepackage{hyperref}
\hypersetup{
    colorlinks=true,
    citecolor=RawSienna,
    urlcolor=black,
    linkcolor=black
}
\usepackage{latexsym}
\usepackage{amsmath}
\usepackage{mathtools}
\usepackage{amsfonts}
\usepackage{mathrsfs}
\usepackage{multirow}
\usepackage{tablefootnote}

\DeclareMathOperator{\sigmoid}{sigmoid}

\newcommand*{\affaddr}[1]{#1} 
\newcommand*{\affmark}[1][*]{\textsuperscript{#1}}
\newcommand*{\email}[1]{\texttt{#1}}

\title{An Empirical Exploration of Skip Connections for Sequential Tagging}

\author{%
Huijia Wu\affmark[1,3], Jiajun Zhang\affmark[1,3], and Chengqing Zong\affmark[1,2,3]\\
\affaddr{\affmark[1]National Laboratory of Pattern Recognition, Institute of Automation, CAS}\\
\affaddr{\affmark[2]CAS Center for Excellence in Brain Science and Intelligence Technology }\\
\affaddr{\affmark[3]University of Chinese Academy of Sciences} \\
\email{\{huijia.wu,jjzhang,cqzong\}@nlpr.ia.ac.cn}
}
\date{}

\begin{document}
\maketitle
\begin{abstract}
  In this paper, we empirically explore the effects of various kinds of skip connections in stacked bidirectional LSTMs for sequential tagging. We investigate three kinds of skip connections connecting to LSTM cells: (a) skip connections to the gates, (b) skip connections to the internal states and (c) skip connections to the cell outputs. We present comprehensive experiments showing that skip connections to cell outputs outperform the remaining two. Furthermore, we observe that using gated identity functions as skip mappings works pretty well. Based on this novel skip connections, we successfully train deep stacked bidirectional LSTM models and obtain state-of-the-art results on CCG supertagging and comparable results on POS tagging.
\end{abstract}

\section{Introduction}
\label{intro}

\blfootnote{
    %
    %
    %
    %
    %
    %
    \hspace{-0.65cm}  
    This work is licensed under a Creative Commons 
    Attribution 4.0 International License.
    License details:
    \url{http://creativecommons.org/licenses/by/4.0/}
}

In natural language processing, sequential tagging mainly refers to the tasks of assigning discrete labels to each token in a sequence. Typical examples include part-of-speech (POS) tagging and combinatory  category grammar (CCG) supertagging. A regular feature of sequential tagging is that the input tokens in a sequence cannot be assumed to be independent since the same token in different contexts can be assigned to different tags. Therefore, the classifier should have memories to remember the contexts to make a correct prediction.

Bidirectional LSTMs \cite{graves2005framewise} become dominant in sequential tagging problems due to the superior performance \cite{wang2015part,vaswani2016supertagging,lample2016neural}. The horizontal hierarchy of LSTMs with bidirectional processing can remember the long-range dependencies without affecting the short-term storage. Although the models have a deep horizontal hierarchy (the depth is the same as the sequence length), the vertical hierarchy is often shallow, which may not be efficient at representing each token. Stacked LSTMs are deep in both directions, but become harder to train due to the feed-forward structure of stacked layers. 

Skip connections (or shortcut connections) enable unimpeded information flow by adding direct connections across different layers \cite{raiko2012deep,graves2013generating,hermans2013training}. However, there is a lack of exploration and analyzing various kinds of skip connections in stacked LSTMs. There are two issues to handle skip connections in stacked LSTMs: One is where to add the skip connections, the other is what kind of skip connections should be used to pass the information. To answer the first question, we empirically analyze three positions of LSTM blocks to receive the previous layer's output. For the second one, we present an identity mapping to receive the previous layer's outputs. Furthermore, following the gate design of LSTM \cite{hochreiter1997lstm,gers2000learning} and highway networks \cite{srivastava2015training,srivastava2015highway}, we observe that adding a multiplicative gate to the identity function will help to improve performance.

In this paper, we present a neural architecture for sequential tagging. The input of the network are token representations. We concatenate word embeddings to character embeddings to represent the word and morphemes. A deep stacked bidirectional LSTM with well-designed skip connections is then used to extract the features needed for classification from the inputs. The output layer uses \emph{softmax} function to output the tag distribution for each token. 

Our main contribution is that we empirically evaluated the effects of various kinds of skip connections within stacked LSTMs. We present comprehensive experiments on the supertagging task showing that skip connections to the cell outputs using identity function multiplied with an exclusive gate can help to improve the network performance. Our model is evaluated on two sequential tagging tasks, obtaining state-of-the-art results on CCG supertagging and comparable results on POS tagging.

\section{Related Work}
Skip connections have been widely used for training deep neural networks. For recurrent neural networks, Schmidhuber \shortcite{schmidhuber1992learning}; El Hihi and Bengio \shortcite{el1995hierarchical} introduced deep RNNs by stacking hidden layers on top of each other. Raiko et al. \shortcite{raiko2012deep}; Graves \shortcite{graves2013generating};  Hermans and Schrauwen \shortcite{hermans2013training} proposed the use of skip connections in stacked RNNs. However, the researchers have paid less attention to the analyzing of various kinds of skip connections, which is our focus in this paper.

The works closely related to ours are Srivastava et al. \shortcite{srivastava2015highway}, He et al. \shortcite{he2015deep}, Kalchbrenner et al. \shortcite{kalchbrenner2015grid}, Yao et al. \shortcite{yao2015depth}, Zhang et al. \shortcite{zhang2016highway}, and Zilly et al. \shortcite{zilly2016recurrent}. These works are all based on the design of extra connections between different layers. Srivastava et al. \shortcite{srivastava2015highway} and He et al. \shortcite{he2015deep} mainly focus on feed-forward neural network, using well-designed skip connections across different layers to make the information pass more easily. The Grid LSTM proposed by Kalchbrenner et al. \shortcite{kalchbrenner2015grid} extends the one dimensional LSTMs to many dimensional LSTMs, which provides a more general framework to construct deep LSTMs. 

Yao et al. \shortcite{yao2015depth} and Zhang et al. \shortcite{zhang2016highway} propose highway LSTMs by introducing gated direct connections between internal states in adjacent layers and do not use skip connections, while we propose gated skip connections across cell outputs. Zilly et al. \shortcite{zilly2016recurrent} introduce recurrent highway networks (RHN) which use a single recurrent layer to make RNN deep in a vertical direction, in contrast to our stacked models.

\section{Recurrent Neural Networks for Sequential Tagging}
Consider a recurrent neural network applied to sequential tagging: Given a sequence $x = (x_1, \ldots, x_T)$, the RNN computes the hidden state $h = (h_1, \ldots, h_T)$ and the output $y = (y_1, \ldots, y_T)$ by iterating the following equations:
\begin{align}
    h_t &= f(x_t, h_{t-1}; \theta_h) \label{rnn_trans} \\
    y_t &= g(h_t; \theta_o)
\end{align}
where $t \in \{1, \ldots, T\}$ represents the time. $x_t$ represents the input at time $t$, $h_{t-1}$ and $h_t$ are the previous and the current hidden state, respectively. $f$ and $g$ are the transition function and the output function, respectively. $\theta_h$ and $\theta_o$ are network parameters.

We use a negative log-likelihood cost to evaluate the performance, which can be written as:
\begin{align}
\mathcal{C} = - \frac{1}{N} \sum_{n=1}^N \log {y}_{t^n}
\end{align}
where $t^n \in \mathbb{N}$ is the true target for sample $n$, and ${y}_{t^n}$ is the $t$-th output in the \textit{softmax} layer given the inputs ${x}^n$.

The core idea of Long Short-Term Memory networks is to replace \eqref{rnn_trans} with the following equation:
\begin{equation}
    c_t = f(x_t, h_{t-1}) + c_{t-1}
\end{equation}
where $c_t$ is the internal state of the memory cell, which is designed to store the information for much longer time. Besides this, LSTM uses gates to avoid weight update conflicts. 

Standard LSTMs process sequences in temporal order, which will ignore future context. Bidirectional LSTMs solve this problem by combining both the forward and the backward processing of the input sequences using two separate recurrent hidden layers:
\begin{align}
    \overrightarrow{h_t} &= \text{LSTM}(\overrightarrow{x_t}, \overrightarrow{h_{t-1}}, \overrightarrow{c_{t-1}}) \\
    \overleftarrow{h_t} &= \text{LSTM}(\overleftarrow{x_t}, \overleftarrow{h_{t-1}}, \overleftarrow{c_{t-1}}) \\
    y_t &= g(\overrightarrow{h_t}, \overleftarrow{h_t})
\end{align}
where $\text{LSTM}(\cdot)$ is the LSTM computation. $\overrightarrow{x_t}$ and $\overleftarrow{x_t}$ are the forward and the backward input sequence, respectively. The output of the two hidden layers $\overrightarrow{h_t}$ and $\overleftarrow{h_t}$ in a birectional LSTM are connected to the output layer.

Stacked RNN is one type of deep RNNs, which refers to the hidden layers are stacked on top of each other, each feeding up to the layer above:
\begin{align}
    h_t^{l} = f^{l}(h_t^{l-1}, h_{t-1}^{l})
\end{align}
where $h_t^{l}$ is the $t$-th hidden state of the $l$-th layer.

\section{Various kinds of Skip Connections} \label{skip}
Skip connections in simple RNNs are trivial since there is only one position to connect to the hidden units. But for stacked LSTMs, the skip connections need to be carefully treated to train the network successfully. In this section, we analyze and compare various types of skip connections. At first, we give a detailed definition of stacked LSTMs, which can help us to describe skip connections. Then we start our construction of skip connections in stacked LSTMs. At last, we formulate various kinds of skip connections. 

Stacked LSTMs without skip connections can be defined as:
\begin{align} \label{no_skip}
    \left(\!
    \begin{array}{c}
      i_t^{l} \\
      f_t^{l} \\
      o_t^{l} \\
      s_t^{l}
    \end{array}
    \!\right) = 
    \left(\!
    \begin{array}{c}
      \text{sigm} \\
      \text{sigm} \\
      \text{sigm} \\
      \text{tanh}
    \end{array}
    \!\right) W^{l} 
    \left(\!
    \begin{array}{c}
    h_t^{l-1} \\
    h_{t-1}^{l}
    \end{array}
    \!\right) & &
    \begin{aligned}
        c_t^l &= f_t^l \odot c_{t-1}^l + i_t^l \odot s_t^{l} \\
        h_t^l &= o_t^l \odot \text{tanh}(c_t^l)
    \end{aligned}
\end{align}
During forward pass, LSTM needs to calculate $c_t^l$ and $h_t^l$, which is the cell's internal state and the cell outputs state, respectively. To get $c_t^l$, $s_t^l$ needs to be computed to store the current input. Then this result is multiplied by the input gate $i_t^{l}$, which decides when to keep or override information in memory cell $c_t^{l}$. The cell is designed to store the previous information $c_{t-1}^l$, which can be reset by a forget gate $f_t^l$. The new cell state is then obtained by adding the result to the current input. The cell outputs $h_t^l$ are computed by multiplying the activated cell state by the output gate $o_t^l$, which learns when to access memory cell and when to block it. ``$\text{sigm}$'' and ``$\text{tanh}$'' are the sigmoid and tanh activation function, respectively. $W^l \in \mathbb{R}^{4n \times 2n}$ is the weight matrix needs to be learned.

The hidden units in stacked LSTMs have two forms. One is the hidden units in the same layer $\{h_t^l, t \in 1, \ldots, T\}$, which are connected through an LSTM. The other is the hidden units at the same time step $\{h_t^{l}, l \in 1, \ldots, L\}$, which are connected through a feed-forward network. LSTM can keep the short-term memory for a long time, thus the error signals can be easily passed through $\{1, \ldots, T\}$. However, when the number of stacked layers is large, the feed-forward network will suffer the gradient vanishing/exploding problems, which make the gradients hard to pass through $\{1, \ldots, L\}$. 

The core idea of LSTM is to use an identity function to make the constant error carrosel. He et al. \shortcite{he2015deep} also use an identity mapping to train a very deep convolution neural network with improved performance. All these inspired us to use an identity function for the skip connections. Rather, the gates of LSTM are essential parts to avoid weight update conflicts, which are also invoked by skip connections. Following highway gating, we use a gate multiplied with identity mapping to avoid the conflicts.

Skip connections are cross-layer connections, which means that the output of layer $l-$2 is not only connected to the layer $l-$1, but also connected to layer $l$. For stacked LSTMs, $h_t^{l-2}$ can be connected to the gates, the internal states, and the cell outputs in layer $l$'s LSTM blocks. We formalize these below:

\subparagraph{Skip connections to the gates.}
We can connect $h_t^{l-2}$ to the gates through an identity mapping:
\begin{align} \label{to_gate}
    \left(\!
    \begin{array}{c}
      i_t^{l} \\
      f_t^{l} \\
      o_t^{l} \\
      s_t^{l} 
    \end{array}
    \!\right) = 
    \left(\!
    \begin{array}{c}
      \text{sigm} \\
      \text{sigm} \\
      \text{sigm} \\
      \text{tanh}
    \end{array}
    \!\right)
    \left(\!
    \begin{array}{c}
         W^{l} \, \color{red}{I^l}
    \end{array}
    \!\right)
    \left(\!
    \begin{array}{c}
    h_t^{l-1} \\
    h_{t-1}^{l} \\
    \color{red}{h_t^{l-2}}
    \end{array}
    \!\right)
\end{align}
where $I^l \in \mathbb{R}^{4n \times n}$ is the identity mapping.

\subparagraph{Skip connections to the internal states.}
Another kind of skip connections is to connect $h_t^{l-2}$ to the cell's internal state $c_t^l$:
\begin{align}
    c_t^l &= f_t^l \odot c_{t-1}^l + i_t^l \odot {s_t^{l}} + \color{red}{h_t^{l-2}} \label{to_internal} \\
    h_t^l &= o_t^l \odot \text{tanh}(c_t^l)
\end{align}

\subparagraph{Skip connections to the cell outputs.}
We can also connect $h_t^{l-2}$ to cell outputs:
\begin{align}
    c_t^l &= f_t^l \odot c_{t-1}^l + i_t^l \odot s_t^{l} \\
    h_t^l &= o_t^l \odot \text{tanh}(c_t^l) + \color{red}{h_t^{l-2}} \label{to_output}
\end{align}

\subparagraph{Skip connections using gates.}
Consider the case of skip connections to the cell outputs. The cell outputs grow linearly during the presentation of network depth, which makes the $h_t^l$'s derivative vanish and hard to convergence. Inspired by the introduction of LSTM gates, we add a gate to control the skip connections through retrieving or blocking them:
\begin{align} \label{output_gate}
\left(\!
\begin{array}{c}
  i_t^{l} \\
  f_t^{l} \\
  o_t^{l} \\
  \color{red}{g_t^{l}} \\
  s_t^{l}
\end{array}
\!\right) = 
\left(\!
\begin{array}{c}
  \text{sigm} \\
  \text{sigm} \\
  \text{sigm} \\
  \text{sigm} \\
  \text{tanh}
\end{array} 
\!\right)
W^{l}
\left(\!
\begin{array}{c}
h_t^{l-1} \\
h_{t-1}^{l} \\
\end{array}
\!\right) & &
\begin{aligned}
    c_t^l &= f_t^l \odot c_{t-1}^l + i_t^l \odot s_t^{l} \\
    h_t^l &= o_t^l \odot \text{tanh}(c_t^l) + \color{red}{g_t^l} \odot \color{red}{h_t^{l-2}}
\end{aligned}
\end{align}
where $g_t^l$ is the gate which can be used to access the skipped output $h_t^{l-2}$ or block it. When $g_t^l$ equals 0, no skipped output can be passed through skip connections, which is equivalent to traditional stacked LSTMs. Otherwise, it behaves like a feed-forward LSTM using gated identity connections. Here we omit the case of adding gates to skip connections to the internal state, which is similar to the above case. 

\subparagraph{Skip connections in bidirectional LSTM.}
Using skip connections in bidirectional LSTM is similar to the one used in LSTM, with a bidirectional processing:
\begin{align}
\begin{aligned}
    \overrightarrow{c_t^l} &= \overrightarrow{f} \odot \overrightarrow{c_{t-1}^l} + \overrightarrow{i} \odot \overrightarrow{s_t^l} \\
    \overrightarrow{h_t^l} &= \overrightarrow{o} \odot \text{tanh}(\overrightarrow{c_t^l}) + \color{red}{\overrightarrow{g}} \odot \color{red}{\overrightarrow{h_t^{l-2}}}
\end{aligned}
    & &
\begin{aligned}
    \overleftarrow{c_t^l} &= \overleftarrow{f} \odot \overleftarrow{c_{t-1}^l} + \overleftarrow{i} \odot \overleftarrow{s_t^l} \\
    \overleftarrow{h_t^l} &= \overleftarrow{o} \odot \text{tanh}(\overleftarrow{c_t^l}) + \color{red}{\overleftarrow{g}} \odot \color{red}{\overleftarrow{h_t^{l-2}}}
\end{aligned}
\end{align}

\section{Neural Architecture for Sequential Tagging}
Sequential tagging can be formulated as $P({t} | {w}; {\theta})$, where ${w} = [w_1, \ldots, w_T]$ indicates the $T$ words in a sentence, and ${t} = [t_1, \ldots, t_T]$ indicates the corresponding $T$ tags. In this section we introduce an neural architecture for $P(\cdot)$, which includes an input layer, a stacked hidden layers and an output layer. Since the stacked hidden layers have already been introduced, we only introduce the input and the output layer here.

\subsection{Network Inputs}
Network inputs are the representation of each token in a sequence. There are many kinds of token representations, such as using a single word embedding, using a local window approach, or a combination of word and character-level representation. Here our inputs contain the concatenation of word representations, character representations, and capitalization representations.

\subparagraph{Word representations.} 
All words in the vocabulary share a common look-up table, which is initialized with random initializations or pre-trained embeddings. Each word in a sentence can be mapped to an embedding vector $w_t$. The whole sentence is then represented by a matrix with columns vector $[w_1, w_2, \ldots, w_T]$. We use a context window of size $d$ surrounding with a word $w_t$ to get its context information. Following Wu et al. \shortcite{wu2016dynamic}, we add logistic gates to each token in the context window. The word representation is computed as $w_t = [r_{{t-\lfloor d/2 \rfloor}} w_{t-\lfloor d/2 \rfloor}; \ldots; r_{{t+\lfloor d/2 \rfloor}} w_{t+\lfloor d/2 \rfloor}]$, where $r_t := [r_{{t-\lfloor d/2 \rfloor}}, \ldots, r_{{t+\lfloor d/2 \rfloor}}] \in \mathbb{R}^{d}$ is a logistic gate to filter the unnecessary contexts, $w_{t-\lfloor d/2 \rfloor}, \ldots, w_{t+\lfloor d/2 \rfloor}$ is the word embeddings in the local window.

\subparagraph{Character representations.} Prefix and suffix information about words are important features in sequential tagging. Inspired by Fonseca et al. \shortcite{fonseca2015evaluating} et al, which uses a character prefix and suffix with length from 1 to 5 for part-of-speech tagging, we concatenate character embeddings in a word to get the character-level representation. Concretely, given a word $w$ consisting of a sequence of characters $[c_1, c_2, \ldots, c_{l_w}]$, where $l_w$ is the length of the word and $L(\cdot)$ is the look-up table for characters. We concatenate the leftmost most 5 character embeddings $L(c_1), \ldots, L(c_5)$ with its rightmost 5 character embeddings $L(c_{l_w-4}), \ldots, L(c_{l_w})$. When a word is less than five characters, we pad the remaining characters with the same special symbol.


\subparagraph{Capitalization representations.}
We lowercase the words to decrease the size of word vocabulary to reduce sparsity, but we need an extra capitalization embeddings to store the capitalization features, which represent whether or not a word is capitalized.

\subsection{Network Outputs}
For sequential tagging, we use a \emph{softmax} activation function $g(\cdot)$ in the output layer:
\begin{align}
{y}_t &= g({W^{hy}} [\overrightarrow{h_t}; \overleftarrow{h_t}])
\end{align}
where ${y}_t$ is a probability distribution over all possible tags. $y_k(t) = \frac{\exp(h_k)}{\sum_{k'} \exp(h_{k'})}$ is the $k$-th dimension of ${y}_t$, which corresponds to the $k$-th tags in the tag set. ${W^{hy}}$ is the hidden-to-output weight.

\section{Experiments}
\subsection{Combinatory Category Grammar Supertagging}
Combinatory Category Grammar (CCG) supertagging is a sequential tagging problem in natural language processing. The task is to assign supertags to each word in a sentence. In CCG the supertags stand for the lexical categories, which are composed of the basic categories such as $N$, $NP$ and $PP$, and complex categories, which are the combination of the basic categories based on a set of rules. Detailed explanations of CCG refers to \cite{steedman2000syntactic,steedman2011combinatory}. 

The training set of this task only contains 39604 sentences, which is too small to train a deep model, and may cause over-parametrization. But we choose it since it has been already proved that a bidirectional recurrent net fits the task by many authors \cite{lewis2016lstm,vaswani2016supertagging}.

\subsubsection{Dataset and Pre-processing}
Our experiments are performed on CCGBank \cite{hockenmaier2007ccgbank}, which is a translation from Penn Treebank \cite{marcus1993building} to CCG with a coverage 99.4\%. We follow the standard splits, using sections 02-21 for training, section 00 for development and section 23 for the test. We use a full category set containing 1285 tags. All digits are mapped into the same digit `9', and all words are lowercased.

\subsubsection{Network Configuration}
\subparagraph{Initialization.}
There are two types of weights in our experiments: recurrent and non-recurrent weights. For non-recurrent weights, we initialize word embeddings with the pre-trained 200-dimensional GolVe vectors \cite{pennington2014glove}. Other weights are initialized with the Gaussian distribution $\mathcal{N}(0, \frac{1}{\sqrt{\text{fan-in}}})$ scaled by a factor of 0.1, where \textit{fan-in} is the number of units in the input layer. For recurrent weight matrices, following Saxe et al. \shortcite{saxe2013exact} we initialize with random orthogonal matrices through SVD to avoid unstable gradients. Orthogonal initialization for recurrent weights is important in our experiments, which takes about $2\%$ relative performance enhancement than other methods such as Xavier initialization \cite{glorot2010understanding}.

\subparagraph{Hyperparameters.}
For the word representations, we use a small window size of 3 for the convolutional layer. The dimension of the word representation after the convolutional operation is 600. The size of character embedding and capitalization embeddings are set to 5. The number of cells of the stacked bidirectional LSTM is set to 512. We also tried 400 cells or 600 cells and found this number did not impact performance so much. All stacked hidden layers have the same number of cells. The output layer has 1286 neurons, which equals to the number of tags in the training set with a \textsc{rare} symbol. 
\subparagraph{Training.} 
We train the networks using the back-propagation algorithm, using stochastic gradient descent (SGD) algorithm with an equal learning rate 0.02 for all layers. We also tried other optimization methods, such as momentum \cite{plaut1986experiments}, Adadelta \cite{zeiler2012adadelta}, or Adam \cite{kingma2014adam}, but none of them perform as well as SGD. Gradient clipping is not used. We use on-line learning in our experiments, which means the parameters will be updated on every training sequences, one at a time. We trained the 7-layer network for roughly 2 to 3 days on one NVIDIA TITAN X GPU using Theano \footnote{\url{http://deeplearning.net/software/theano/}} \cite{Team2016Theano}.

\subparagraph{Regularization.}
Dropout \cite{srivastava2014dropout} is the only regularizer in our model to avoid overfitting. Other regularization methods such as weight decay and batch normalization do not work in our experiments. We add a binary dropout mask to the local context windows on the embedding layer, with a drop rate $p$ of 0.25. We also apply dropout to the output of the first hidden layer and the last hidden layer, with a 0.5 drop rate. At test time, weights are scaled with a factor $1-p$. 
\subsubsection{Results}
Table \ref{CCGBank} shows the comparisons with other models for supertagging. The comparisons do not include any externally labeled data and POS labels. We use stacked bidirectional LSTMs with gated skip connections for the comparisons, and report the highest 1-best supertagging accuracy on the development set for final testing.
Our model presents state-of-the-art results compared to the existing systems. The character-level information (+ 3\% relative accuracy) and dropout (+ 8\% relative accuracy) are necessary to improve the performance.

\begin{table*}[t]
\begin{center}
\begin{tabular}{ l|l|l }
\hline \bf Model & \bf Dev & \bf Test \\ \hline \hline
Clark and Curran \shortcite{clark2007wide} & 91.5 & 92.0 \\
Lewis et al. \shortcite{lewis2014improved}  & 91.3 & 91.6 \\
Lewis et al. \shortcite{lewis2016lstm} & 94.1 & 94.3 \\
Xu et al. \shortcite{xu2015ccg} & 93.1 & 93.0 \\
Xu et al. \shortcite{xu2016expected} & 93.49 & 93.52 \\
Vaswani et al. \shortcite{vaswani2016supertagging} & 94.24 & 94.5 \\ \hline
7-layers + skip output + gating & 94.51 & 94.67 \\
7-layers + skip output + gating (no char) & 94.33 & 94.45 \\
7-layers + skip output + gating (no dropout) & 94.06 & 94.0 \\
9-layers + skip output + gating & \bf 94.55 & \bf 94.69 \\
\hline
\end{tabular}
\end{center}
\caption{\label{CCGBank} 1-best supertagging accuracy on CCGbank. ``skip output'' refers to the skip connections to the cell output, ``gating'' refers to adding a gate to the identity function, ``no char'' refers to the models that do not use the character-level information, ``no dropout'' refers to models that do not use dropout.}
\end{table*}

\subsubsection{Experiments on Skip Connections}
We experiment with a 7-layer model on CCGbank to compare different kinds of skip connections introduced in Section \ref{skip}. Our analysis mainly focuses on the identity function and the gating mechanism. The comparisons (Table \ref{exp_skip}) are summarized as follows:
\subparagraph{No skip connections.} When the number of stacked layers is large, the performance will degrade without skip connections. The accuracy in a 7-layer stacked model without skip connections is 93.94\% (Table \ref{exp_skip}), which is lower than the one using skip connections.

\subparagraph{Various kinds of skip connections.} We experiment with the gated identity connections between internal states introduced in Zhang et al.\shortcite{zhang2016highway}, but the network performs not good (Table \ref{exp_skip}, 93.14\%). We also implement the method proposed in Zilly et al. \shortcite{zilly2016recurrent}, which we use a single bidirectional RNH layer with a recurrent depth of 3 with a slightly modification\footnote{Our original implementation of Zilly et a. \shortcite{zilly2016recurrent} with a recurrent depth of 3 fails to converge. The reason might be due to the explosion of $s_{L}^t$ under addition. To avoid this, we replace $s_{L}^t$ with $o_t * \tanh(s_{L}^t)$ in the last recurrent step.}. Skip connections to the cell outputs with identity function and multiplicative gating achieves the highest accuracy (Table \ref{exp_skip}, 94.51\%) on the development set. We also observe that skip to the internal states without gate get a slightly better performance (Table \ref{exp_skip}, 94.33\%) than the one with gate (94.24\%) on the development set. Here we recommend to set the forget bias to 0 to get a better development accuracy.

\subparagraph{Identity mapping.}
We use the $\sigmoid$ function to the previous outputs to break the identity link, in which we replace $g_t \odot h_t^{l-1}$ in Eq. \eqref{output_gate} with $g_t \odot \sigma(h_t^{l-1})$, where $\sigma(x) = \frac{1}{1 + e^{-x}}$. The result of the sigmoid function is 94.02\% (Table \ref{exp_skip}), which is poor than the identity function. We can infer that the identity function is more suitable than other scaled functions such as sigmoid or tanh to transmit information.

\subparagraph{Exclusive gating.}
Following the gating mechanism adopted in highway networks, we consider adding a gate $g_t$ to make a flexible control to the skip connections. Our gating function is $g_t^{l} = \sigma(W_g^l h_{t-1}^l + U_g^l h_t^{l-2})$. Gated identity connections are essential to achieving state-of-the-art result on CCGbank.

\begin{table*}[!htbp]
\begin{center}
\begin{tabular}{ l|l|l|l}
\hline 
\bf Case & \bf Variant & \bf Dev & \bf Test \\ 
\hline \hline
H-LSTM, Zhang et al.\shortcite{zhang2016highway} & - & 93.14 & 93.52 \\ 
\hline
RHN, Zilly et al. \shortcite{zilly2016recurrent} & $L=3$, with output gates & 94.28 & 94.24 \\
\hline
no skip connections & - & 93.94 & 94.26 \\
\hline
\multirow{1}{*}{to the gates, Eq. \eqref{to_gate}} 
& - & 93.9 & 94.22 \\ 
\hline
\multirow{2}{*}{to the internals} 
& no gate, Eq. \eqref{to_internal} & 94.33 & 94.63 \\ 
& with gate & 94.24 & 94.52 \\
\hline
\multirow{4}{*}{to the outputs}
& no gate, Eq. \eqref{to_output} & 93.89 & 93.98 \\ 
& with gate, $b_f=5$, Eq. \eqref{output_gate} & 94.23 & 94.81 \\ 
& with gate, $b_f=0$, Eq. \eqref{output_gate} & 94.51 & 94.67 \\ 
& $\sigmoid$ mapping: $g_t \odot \sigma(h_t^{l-1})$ & 94.02 & 94.18 \\
\hline
\end{tabular}
\end{center}
\caption{\label{exp_skip} Accuracy on CCGbank using 7-layer stacked bidirectional LSTMs, with different types of skip connections. $b_f$ is the bias of the forget gate. We report ``fail'' when the validation error is higher than 10\%.}
\end{table*}

\subsubsection{Experiments on Number of Layers}
Table \ref{exp_layer} compares the effect of the depth in the stacked models. We can observe that the performance is getting better with the increased number of layers. But when the number of layers exceeds 9, the performance will be hurt. In the experiments, we found that the number of stacked layers between 7 and 9 are the best choice using skip connections. Notice that we do not use layer-wise pretraining \cite{bengio2007greedy,simonyan2014very}, which is an important technique in training deep networks. Further improvements might be obtained with this method to build a deeper network with improved performance.

\begin{table*}[!htbp]
\begin{center}
\begin{tabular}{ l|l|l}
\hline 
\bf \#Layers & \bf Dev & \bf Test \\ 
\hline \hline
\multirow{1}{*}{3} 
& 94.21 & 94.35 \\ 
\multirow{1}{*}{5} 
& 94.51 & 94.57 \\
\multirow{1}{*}{7}
& 94.51 & 94.67 \\ 
\multirow{1}{*}{9}
& 94.55 & 94.7 \\ 
\multirow{1}{*}{11}
& 94.43 & 94.65 \\ 
\hline
\end{tabular}
\end{center}
\caption{\label{exp_layer} Accuracy on CCGbank using gated identity connections to cell outputs, with different number of stacked layers.}
\end{table*}

\subsection{Part-of-speech Tagging}
Part-of-speech tagging is another sequential tagging task, which is to assign POS tags to each word in a sentence. It is very similar to the supertagging task. Therefore, these two tasks can be solved in a unified architecture. For POS tagging, we use the same network configurations as supertagging, except the word vocabulary size and the tag set size. We conduct experiments on the Wall Street Journal of the Penn Treebank dataset, adopting the standard splits (sections 0-18 for the train, sections 19-21 for validation and sections 22-24 for testing). 

\begin{table*}[!htbp]
\begin{center}
\begin{tabular}{ l|l}
\hline \bf Model & \bf Test \\ \hline \hline
S\o gaard \shortcite{sogaard2011semisupervised} & 97.5 \\
Ling et al. \shortcite{ling2015finding}  & 97.36 \\
Wang et al. \shortcite{wang2015part} & \bf 97.78 \\
Vaswani et al. \shortcite{vaswani2016supertagging} & 97.4 \\ \hline
7-layers + skip output + gating & 97.45 \\
9-layers + skip output + gating & 97.45 \\
\hline
\end{tabular}
\end{center}
\caption{\label{pos} Accuracy for POS tagging on WSJ.}
\end{table*}

Although the POS tagging result presented in Table \ref{pos} is slightly below the state-of-the-art, we neither do any parameter tunings nor change the network architectures, just use the one getting the best development accuracy on the supertagging task. This proves the generalization of the model and avoids heavy work of model re-designing.

\section{Conclusions}
This paper investigates various kinds of skip connections in stacked bidirectional LSTM models. We present a deep stacked network (7 or 9 layers) that can be easily trained and get improved accuracy on CCG supertagging and POS tagging. Our experiments show that skip connections to the cell outputs with the gated identity function performs the best. Our explorations could easily be applied to other sequential processing problems, which can be modelled with RNN architectures.

\section*{Acknowledgements}
The research work has been funded by the Natural Science Foundation of China under Grant No. 61333018. We thank the anonymous reviewers for their useful comments that greatly improved the manuscript.

\bibliographystyle{acl}
\bibliography{coling2016}

\end{document}